\pdfoutput=1

\documentclass[11pt]{article}

\usepackage[]{acl}

\usepackage{times}
\usepackage{latexsym}

\usepackage[T1]{fontenc}

\usepackage[utf8]{inputenc}

\usepackage{microtype}

\usepackage{graphicx}
\usepackage{hyperref}       
\usepackage{url}            
\usepackage{booktabs}       
\usepackage{amsfonts}       
\usepackage{nicefrac}       
\usepackage{xcolor}         

\usepackage{float}
\usepackage{multirow}
\usepackage[normalem]{ulem}
\usepackage{subfigure}

\useunder{\uline}{\ul}{}
\usepackage{MnSymbol,wasysym}
\definecolor{lightblue}{RGB}{2,255,255}
\definecolor{lemonyellow}{RGB}{255,255,0}
\usepackage{caption}

\title{
Premise-based Multimodal Reasoning: Conditional Inference \\ on Joint Textual and Visual Clues
}

\author{
Qingxiu Dong\textsuperscript{\rm1 $*$}, Ziwei Qin\textsuperscript{\rm1 $*$}, Heming Xia\textsuperscript{\rm 1}, Tian Feng\textsuperscript{\rm 1},\\ \textbf{Shoujie Tong}\textsuperscript{\rm 1}, 
\textbf{Haoran Meng}\textsuperscript{\rm 1},
\textbf{Lin Xu}\textsuperscript{\rm 1}, \textbf{Zhongyu Wei}\textsuperscript{\rm 2},\\ \textbf{Weidong Zhan}\textsuperscript{\rm 1}, 
\textbf{Baobao Chang}\textsuperscript{\rm 1},
\textbf{Sujian Li}\textsuperscript{\rm 1}, \textbf{Tianyu Liu}\textsuperscript{\rm 3}, \textbf{Zuifang Sui}\textsuperscript{\rm 1} \\
\textsuperscript{\rm 1} Key Laboratory of Computational Linguistics, Peking University, MOE, China \\
  \textsuperscript{\rm 2} School of Data Science, Fudan University \ \ \textsuperscript{\rm 3} Tencent Cloud Xiaowei \\
  \texttt{ \{dqx,qinziwei\}@stu.pku.edu.cn}
}

\begin{document}
\maketitle
\renewcommand{\thefootnote}{\fnsymbol{footnote}}
\footnotetext[1]{Equal contribution.}
\renewcommand{\thefootnote}{\arabic{footnote}}
\begin{abstract}

It is a common practice for recent works in vision language cross-modal reasoning to adopt a binary or multi-choice classification formulation taking as input a set of source image(s) and textual query. In this work, we take a sober look at such an ``unconditional'' formulation in the sense that no prior knowledge is specified with respect to the source image(s). Inspired by the designs of both visual commonsense reasoning and natural language inference tasks, we propose a new task termed ``\underline{\textbf{P}}remise-based \underline{\textbf{M}}ulti-modal \underline{\textbf{R}}easoning'' (PMR) where a textual premise is the background presumption on each source image.
The PMR dataset contains 15,360 manually annotated samples which are created by a multi-phase crowd-sourcing process. 
With selected high-quality movie screenshots and human-curated premise templates from 6 pre-defined categories, we ask crowd-source workers to write one true hypothesis and three distractors (4 choices) given the premise and image through a cross-check procedure. 
Besides, we generate adversarial samples to alleviate the annotation artifacts and double the size of PMR.
We benchmark various state-of-the-art (pretrained) multi-modal inference models on PMR and conduct comprehensive experimental analyses to showcase the utility of our dataset.
\end{abstract}

\section{Introduction}
Cross-modal reasoning between image and text has been recognized as a fundamental and long-standing task in both academia and industry, which has recently attracted intensive attention from both natural language processing and computer vision communities \citep{VL-BERT,ernie-vil,chen2020uniter}. Researchers try to teach machines to perceive, understand and reason with both visual and textual clues, which mimics human cognitive process \citep{lake2016}.

The canonical form for the cross-modal reasoning tasks usually take the source image(s) as the input and request the inference model to perform a multi-choice classification according to the specified textual query, including visual question answering \citep{vqa1,vqa2}, visual commonsense reasoning \citep{vcr}, visual entailment \citep{snli-ve} and image-text grounding \citep{nlvr1,nlvr2}, etc.
We argue the default cross-modal setting is ``static'' or ``unconditional'' in the sense that no prior presumption or belief is attached to the input images \citep{ren2021iais}. The lack of prior presumption would be insignificant while answering factoid questions according to the images, e.g. ``how many dogs are lying on the grass?'' \citep{nlvr2}  or ``is the bowl to the right of the green apple?'' \citep{hudson2019gqa}. However it may not be the case for more sophisticated cross-modal reasoning that involves human-like cognition and commonsense, e.g. ``what may happen to [personA]?'', ``what will [personA] say to [personB]'' or ``what will [personA] do to [personB]''. The answers to these questions are ambiguous by merely looking at the source images, instead we offer an extra textual premise, which serves as the prior belief to the source images, such as ``to [personA], nothing could be more somber'', ``[personA] is the son of [personB]''. We name the proposed task as ``\underline{\textbf{P}}remise-based \underline{\textbf{M}}ulti-modal \underline{\textbf{R}}easoning'' (PMR)\footnote{The dataset and baseline models can be found in \url{https://2030nlp.github.io/PMR/}.}. In PMR, the inference model should be able to reason with both textual (from the premise) and visual (from images) clues.

\begin{figure*}[!ht] 
\flushleft 
\center{\includegraphics[width=0.9\linewidth]  {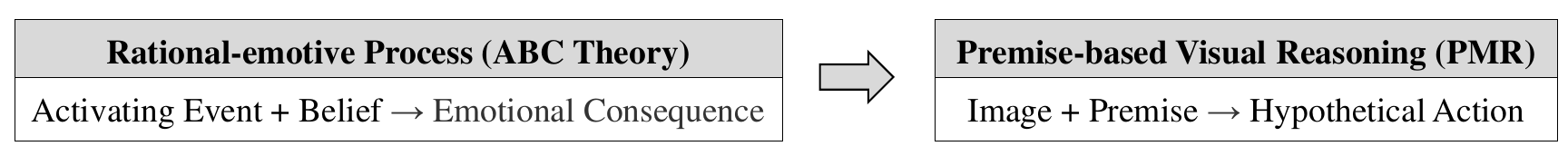}} 
\caption{Connections between the ABC theory \citep{abc} and the proposed PMR.}\label{ABC} 
\end{figure*} 

The motivations for PMR are two-folds. 1) From the social psychology perspective, the design of PMR is inspired by the ABC Theory \citep{abc} , which represents a widely accepted framework for how one's feelings and behavioral patterns are connected. As illustrated in Figure \ref{ABC}, the theory claims that human emotions and resulting behavior do not come directly from the events, but from the interpretations (textual premise) we make of those events (depicted in the source images). 2) From the methodology perspective, PMR gains insights from both visual commonsense reasoning \citep{vcr,lei2020more,park2020visualcomet} and natural language inference \citep{dagan,bowman2019deep,snli-ve} tasks. From the view of commonsense reasoning, the textual premise serves as the prior belief to the image as mentioned before. Meanwhile in the world of natural language inference, the input image can be viewed as the supplementary evidence that supports the textual premise-hypothesis classification. The proposed PMR could be readily seen as a meaningful extension on the joint cross-modal entailment and commonsense reasoning with both visual and textual clues. 

Given the input premise-image pair, the practitioners are requested to choose the only true hypothetical action from four candidates in PMR. All four hypothetical actions are written by crowd-sourced workers in a multi-phase cross-check annotation pipeline. We encourage the annotators to write challenging distracting actions that are logically correct with respect to either the premise or the image, but are contradicted with the joint premise-image pair while combining the visual and textual clues. In order to succeed on PMR, the reasoning model should excel at both language grounding among premise, image and candidate actions, and cross-modal understanding on commonsense and logical inference. We establish multiple competitive cross-modal pretrained models as baselines. We hope the proposed PMR could pave the way for the ``conditional'' cross-modal commonsense and logical reasoning that requires a human-like cognitive process.

\begin{figure*}[!ht] 
\center{\includegraphics[width=\linewidth]  {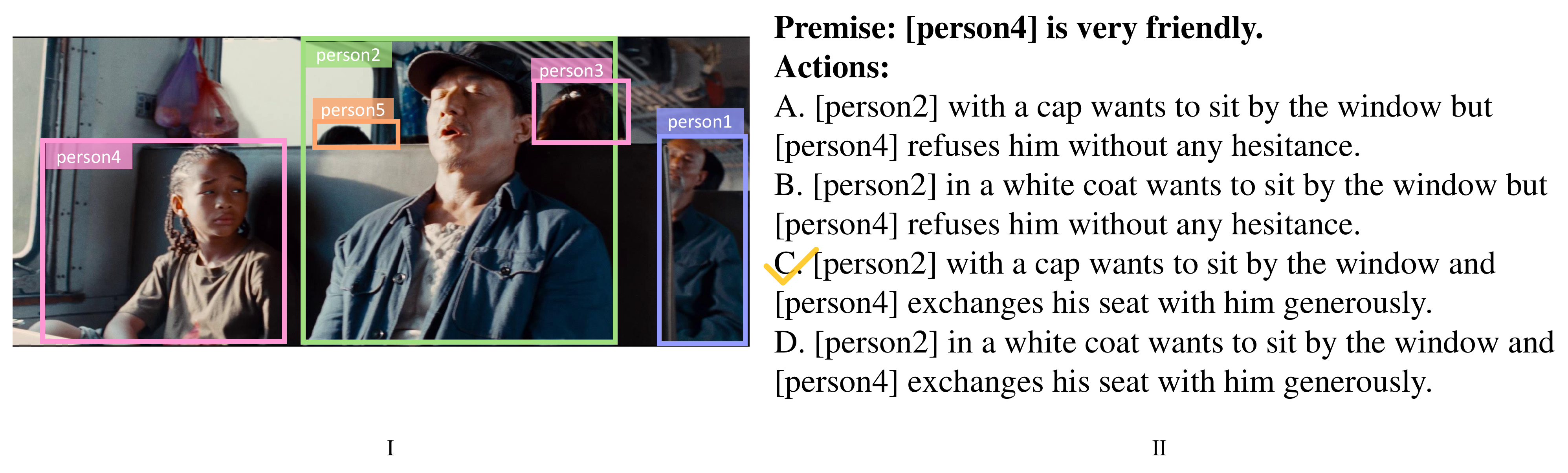}} 
\caption{We demonstrate a source image with object anchors (e.g. `[person1]',`[person2]') in \uppercase\expandafter{\romannumeral1}, and the corresponding premise and hypothetical actions in \uppercase\expandafter{\romannumeral2}.}
\label{mulaction} 
\end{figure*} 

\section{Dataset and Task Overview}
\label{task-description}

We describe the proposed PMR task with an example in Figure \ref{mulaction} \uppercase\expandafter{\romannumeral1}\&\uppercase\expandafter{\romannumeral2}. Given a source image and a textual premise, the inference model should perceive and understand the image in combination with the premise so as to choose the exclusive correct action among the four hypothetical candidates. The premise would serve as the background knowledge or domain-specific commonsense for the given image.
In the running example, the model should be able to recognize what [person2] wears from the image and infer whether [person4] would give his seat to [person2] under the premise ``[person4] is very friendly''. The corrected answer is `C' according to the visual and textual clues.
In total, we collect about 15k instances for PMR. We list the statistics for PMR in Table \ref{tab:data_stats}.

\begin{table*}[!ht]
\centering
\setlength{\tabcolsep}{12pt}
\small
\begin{tabular}{lrrrrrrrr}
\toprule
      & \multicolumn{3}{c}{\textbf{Ori.}} & \multicolumn{3}{c}{\textbf{Adv.}} & \multirow{3}{*}{\textbf{Total~}} \\
      \cmidrule(lr){2-4}\cmidrule(lr){5-7}
      & \textbf{Train} & \textbf{Val}  & \textbf{Test} & \textbf{Train} & \textbf{Val} & \textbf{Test} & \\ \midrule
\#samples                & 12,080 & 1,538 & 1,742  & 12,080  & 1,538 & 1,742 & 30,720\\
\midrule
\#unique 1-gram          & 9,882  & 3,819 & 4,101  & 8,046 & 3,071 & 3,359 & 11,041\\
\#unique 2-gram          & 72,048 & 17,678 & 19,292 & 50,526 & 12,236 & 13,453 & 84,365\\ \midrule
Avg premise length      & 9.48  & 9.47 & 9.54  & 9.48   & 9.47 & 9.54 & 9.49\\
Avg action text length  & 14.38 & 14.41 & 14.45 & 14.20 & 14.42 & 14.31 & 14.31\\
Avg \#objects mentioned & 1.92  & 1.91 & 1.94  & 2.42  & 2.43 & 2.38 &  2.17\\ \midrule
\#images                 & 9,536  & 1,213 & 1,370  & 9,536 & 1,213 & 1,370 & 12,119\\
\#movies covered         & 1,353  & 209  & 170   & 1,353  & 209 & 170 & 1,732\\
\bottomrule
\end{tabular}
\caption{The statistics of PMR dataset. Ori. stands for the manually annotated part of PMR, and Adv. represents the adversarial samples generated automatically.}\label{tab:statistics}
\end{table*}\label{tab:data_stats}

\section{Data Collection}
\label{data-collection}

We collect PMR dataset in a multi-step crowd-sourcing pipeline, including 1) image and premise creation, 2) annotator recruitment and instruction, 3) cross-checking annotation.

\subsection{Image and Premise Creation}
The source images of PMR are selected from the image pool\footnote{The images originate from Fandango MovieClips (\url{https://youtube.com/user/movieclips}) and Large Scale Movie Description Challenge \citep{rohrbach2017movie}.} in the VCR \citep{vcr} dataset. The entire image pool contains 110k high-quality movie screenshots. To make the images fit better with PMR, we screened out those which have low brightness, more than five people, or more than 15 tags, and finally got 29,987 images. We also kept the object anchor information in VCR along with the source images, which was identified using  Mask-RCNN \citep{he2017mask}. 

As there are so many possible choices while writing premise for an image, to facilitate analysis of model performance by premise type, we constrained the premise to six categories and manually wrote templates for each category (see Appendix \ref{premise-templates}). As Figure \ref{fig:premise-classification} shows, the six categories are personality, identity, emotion, relationship, environment and antecedent. To complete templates, we create multiple word sets and fill corresponding words into the slots randomly. And finally, we obtained 30,759 predefined premises, which would be presented to annotators for selection.

\subsection{Annotator Recruitment and Instruction}
Through job descriptions posted on the online forums, we invite more than a hundred applicants who hold a bachelor of arts or higher degree to attend the online pre-annotating instruction and qualification test. In the pre-annotating instruction, we organized a two-hour training session, which covers the basics and goals of PMR, to better instruct the annotators, we extracted the text from VCR as the reference actions. Specifically, we filtered out questions about the following behavior of persons and retrieved the correct answers as the reference actions.

\begin{figure*}[!ht]
    \centering
    \includegraphics[width=\linewidth]{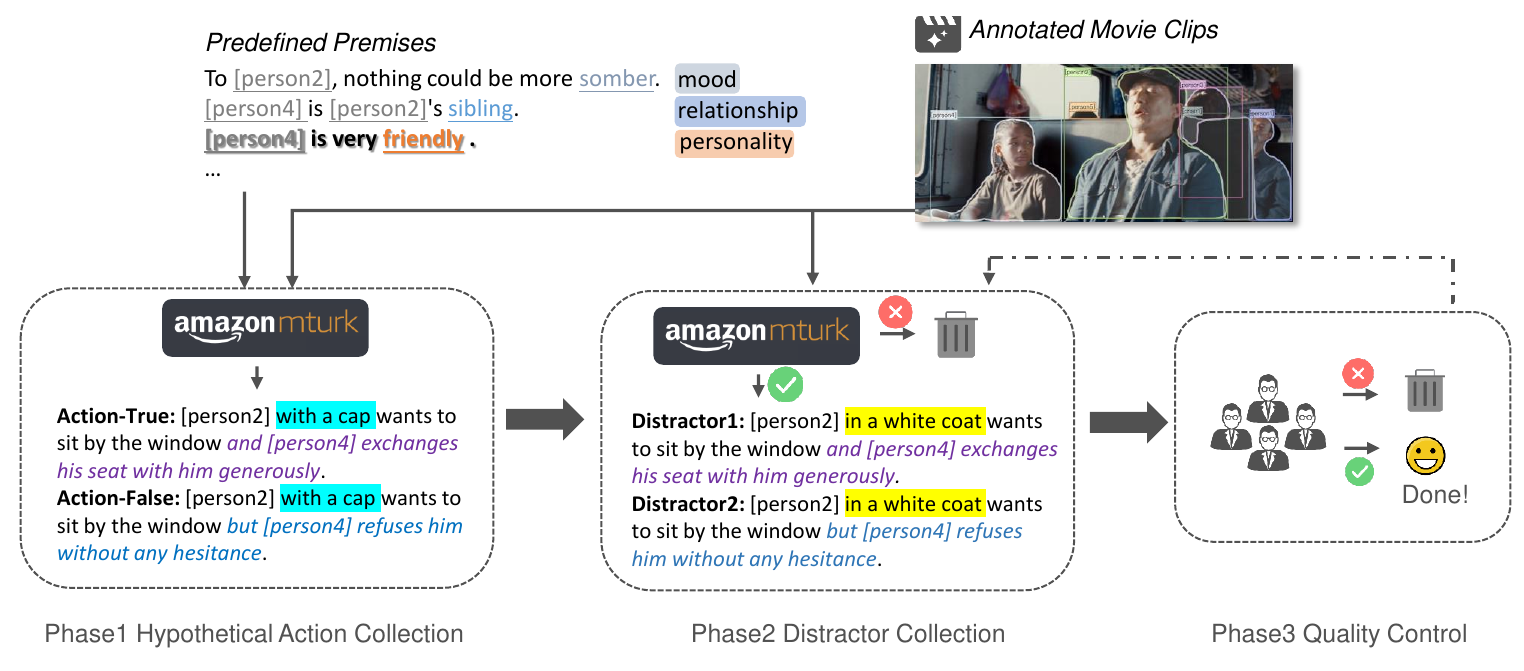}
    \caption{The overview for the cross-check annotation for PMR. In the textual premises, the \uline{underlined} tokens denote the slots in the pre-defined templates. In the hypothetical action collection, the words related with visual information are \colorbox{lightblue}{highlighted in blue}, while the correspondent substitution in the distractors is \colorbox{lemonyellow}{highlighted in yellow}.}
    \label{fig:data-collection}	
\end{figure*}

After the pre-annotating instruction, we conducted a qualification test by asking each participant to write the true and false hypothetical actions for 10 sampled image-premise pairs, the authors manually assess the annotation quality and eventually select 61 annotators whose educational background covers literature, linguistics, sociology, etc. Among them, 19 annotators who got higher scores in the qualification test were recognized as advanced workers and assigned to the distractor collection phase (phase 2 in Figure \ref{fig:data-collection}), while the other annotators were assigned to the hypothetical action creation phase (phase 1 in Figure \ref{fig:data-collection}).

\subsection{Cross-check Annotation}
We divide the annotation process into 3 phases.
In phase 1, workers are presented with an image with bounding boxes on it, six predefined premises of different categories and six reference actions. Firstly, they are supposed to choose an appropriate premise. Workers can adjust the person tags to meet with the given image, but they have to assure that the modified premise still belongs to one of the six categories. Secondly, they ought to write two hypothetical actions, which describe what will happen next. Among them, \textbf{Action-True} contains image information and meets the chosen premise. In contrast, \textbf{Action-False} contains image information but does not meet the chosen premise. Thirdly, to make it easier for replacing certain words in phase 2, they need to enclose the words with curly brackets that mention the information of the image in both written actions.

In phase 2, advanced workers, serving as examiners, are responsible for checking whether the annotation in phase 1 conforms to labeling instruction. They can drop it while meeting poor-quality one, and once they accept it, they need to write another two distractors corresponding to Action-True and Action-False respectively. 
We instruct the annotators to write challenging distractors that are logically correct with respect to either the premise or the image, but contradict with the joint premise-image pair while combining the visual and textual clues.

In phase 3, to ensure the quality of the examiners' work, we sample 10\% HITs in phase 2 to check if the annotation meets our standard. This work is performed by the authors of this paper, and we feedback to examiners timely. 

\paragraph{Post-Processing}
In order to ground objects from images with entities in text, we follow the VCR which substitutes all the tags both in premise and answers with the index in objects list. Besides, since crowd workers can conditionally modify the given premises, which results in the missing of labels of category, we calculate the BLEU \citep{papineni2002bleu} score between premise and each template from six categories and labeled it with the most likely category with the highest BLEU score.

\paragraph{Annotation Cost}
We drop all the annotations that are rejected in phase3 and obtained 15,360 items after 60-day work. (See Appendix \ref{pricing-strategy} for detailed pricing strategy) We split the total dataset into train, valid and test set in 8:1:1.

\paragraph{Adversarial Samples Generation}

To alleviate the possible bias introduced by annotation artifacts, inspired by \citeauthor{vcr}, for each sample in the crowd-sourcing set of PMR, we pick three actions from other samples but similar to the premise as negatives and generate adversarial samples with the same scale. See Appendix \ref{apd-adv} for detailed methods.

\section{Experiments}

To set up the benchmark, we introduce multiple text-only and pretrained multi-modal baseline models as well as human performance on PMR.

\subsection{Baselines}
\paragraph{Text-only Baselines}
We begin with a ``blind'' setting, where only the text is given without access to the image or bounding boxes annotations. In this setting, the models have to choose the correct answers with only textual clues. Specifically we finetune the BERT \citep{Jacob2018bert} model in the action-only and premise+action scenarios. (See Appendix \ref{experimental-details} for training details)

\paragraph{Multi-modal Baselines} Throughout the visual reasoning tasks, cross-modal pretrained models have achieved state-of-the-art performance \citep{vcr,vqa2,hudson2019gqa}. We introduce three powerful pretrained models and test their performance on PMR:

\begin{itemize}
    \item \textbf{VL-BERT}\citep{VL-BERT} A dual-stream pre-trained model, is extended from BERT by appending visual feature embedding along with the subsequent sentences.
    \item \textbf{ERNIE-VL}\citep{ernie-vil} is a knowledge-enhanced approach to learning joint representations of vision and language, which introduced structure knowledge with scene graph prediction tasks while pretraining.
    \item \textbf{UNITER}\citep{chen2020uniter} is also an extension of BERT to the visual domain. It is trained with a conditional masking strategy that allows the model to learn an informative representation of one modality conditioned on the other. 
\end{itemize}

\subsection{Detailed Experiment Settings}

Before feeding into BERT, we concatenate the question and each answer as a sequence and replace detection tags with object names in it. As for person detection tags, we substitute them with gender-neutral names to bridge the gap between PMR and pretraining corpus \citep{vcr}.

In terms of multimodal baselines, while an annotation is processed as above, we regard region features as visual tokens to be concatenated along with text sequence. Region features are extracted with Faster RCNN. Different from ERNIE-VL and UNITER which freeze Faster RCNN all the time, VL-BERT updates it while pretraining and finetuning.

\subsection{Results}
\paragraph{Are Premises Critical for Models to Predict?}As is shown in Table \ref{table:main-result}, E-L achieves the highest accuracy while training and testing on the original split, but if trained with only actions as text inputs, the accuracy of it decreases sharply by 23.3\%.
More details can be found in Table \ref{table:detailed-test}. Premises help E-L correct the prediction of Action-False with the ratio from 30.5\% to 8.2\%.

\begin{table}[!ht]
\small
\centering
\setlength{\tabcolsep}{5.5pt}
\begin{tabular}{@{}lrrrrrrr@{}}
\toprule
                        & \multicolumn{2}{c}{\textbf{Ori. Train}}            & \multicolumn{2}{c}{\textbf{Adv. Train}}            & \multicolumn{3}{c}{\textbf{Mix Train}}    \\ 
                        \cmidrule(l){2-3}\cmidrule(l){4-5}\cmidrule(l){6-8}
                        & \textbf{OT} & \textbf{AT} & \textbf{OT} & \textbf{AT} & \textbf{OT} & \textbf{AT} & \textbf{MT} \\ \midrule
Random               & 25.0     & 25.0               & 
25.0      & 25.0     & 25.0     & 25.0  & 25.0\\ \midrule
B-B$ ^{\dag}$     & 39.3      & 23.1                           & 22.9      & 41.9                           & 25.6      & 25.0      & 25.3       \\
B-B               & 65.2      & 21.8                           & 27.2      & 23.2                           & 25.8      & 26.9      & 26.4       
\\ \midrule
V-B            & 75.4      & 37.4                           & 22.8      & 80.2                           & 70.7      & 66.4      & 68.6       \\
E-B          & 79.0      & 46.2                           & 33.7      & 82.9                           & 72.7      & 76.0      & 74.4       \\
U-B             & 77.4      & 50.7                           & 35.8      & 80.3                           & 72.7      & 70.1      & 73.2      \\ \midrule
V-L           & 79.3      & 47.0                           & 25.3      & 82.5                           & 77.3      & 75.4      & 76.4       \\
E-L$ ^{\dag}$ 
                        & 56.6      & 51.1                           & 40.7
    & 75.5                           & 50.2      & 70.9      & 60.6       \\
E-L         & 79.9      & 52.1                           & 33.4      & 83.6                           & 77.1      & 78.0      & 77.6       \\
U-L            & 77.0      & 57.9                           & 35.7      & 81.9                           & 74.6      & 72.0      & 73.2       \\ \bottomrule
\end{tabular}
\caption{The performance of baselines trained and tested on different dataset split. Ori. Train, Adv. Train and Mix Train respectively stand for the models being trained with the manually annotated part of PMR, adversarial samples and the mixture of the above two. Accordingly, OT, AT and MT represent testing on original, adversarial and mixed test set. For baselines, B, V, E and U are short for BERT, VL-BERT, ERNIE-VIL and UNITER. ``-B'' and ``-L'' stand for the two different sizes of models, ``Base'' and ``Large''. Besides, we trained and tested models without premises as input, which indicated with sign $ ^{\dag}$.
} \label{table:main-result}
\end{table}

\begin{table}[!ht]
\centering
\small
\setlength{\tabcolsep}{4pt}
\begin{tabular}{@{}lrrrrrrrr@{}}
\toprule
     & \multicolumn{4}{c}{\textbf{Ori. Train}} & \multicolumn{4}{c}{\textbf{Mix Train}} \\ 
     \cmidrule(l){2-5}\cmidrule(l){6-9}
     & \textbf{AT}     & \textbf{D1}    & \textbf{AF}    & \textbf{D2}    & \textbf{AT}    & \textbf{D1}    & \textbf{AF}    & \textbf{D2}    \\ \midrule
BERT$ ^{\dag}$ & 39.3   & 15.2  & 31.2  & 14.4  & 25.6  & 23.5  & 26.1  & 24.8  \\
BERT & 65.2   & 19.8  & 10.6  & 4.5   & 25.8  & 25.8  & 24.3  & 24.1  \\
E-L$ ^{\dag}$  & 56.6   & 8.0   & 30.5  & 4.9   & 50.2  & 6.5   & 37.4  & 6.0   \\
E-L  & 79.9   & 10.7  & 8.2   & 1.2   & 77.1  & 9.6   & 11.1  & 2.1   \\ \bottomrule
\end{tabular}
\caption{Detailed performance on Ori. Test. AT, D1, AF and D2 represent the ratio of the four-type prediction of models, Action-True, Distractor1, Action-False and Distractor2 respectively.} \label{table:detailed-test}
\end{table}
\paragraph{Is Vision Modality Useful?}
As for the performance of text-only models and multi-modal models in Table \ref{table:main-result}, considering the information from images improves the accuracy by 10\%-14.7\%. Besides, Distractor1, as generated by substituting objects in Action-True, contradicts with corresponding images, which can be hard negatives while the visual signal is imperceptible. In Table \ref{table:detailed-test}, trained on the original split, E-L predicts Distractor1 with a ratio of 10.7\%, nearly half of the prediction of BERT, suggesting that vision modality is beneficial to distinguish distractors.

\paragraph{Does Crowd-sourcing Introduce Annotation Artifacts to Dataset?}
Table \ref{table:main-result} shows that BERT with only actions outperforms random baseline by 14.3 on accuracy, indicating that textual bias is introduced by manual annotations, which is inevitable due to fixed patterns of thinking during labeling but is tolerable. Adding premises as background presumption, BERT can discover the relation between premises and actions with the performance improved a lot. 

We also conduct a quantitative analysis of bias in data. To check whether labels are significantly associated with certain words, we compute the point-wise mutual information (PMI) \cite{Gururangan2018}. As each sample in PMR has four choices, we consider the premise concatenated respectively with four answers as four binary-class items with label \textit{False} if the answer is incorrect, and compute PMI between words and \textit{False} label.

As Table \ref{table:pmi} shows, among the top 30 words, green and red are the top two with the highest PMI, they are commonly used by crowd workers to modify the objects and generate distractors mismatched with images. Other words such as leave, fight and hit,
crowd workers tend to write universal negative containing these words so as to pass the qualification check, which also results in high PMI of these words.
\begin{table}[!t]
\centering
\small
\setlength{\tabcolsep}{4.5pt}
\begin{tabular}{@{}lcccccc@{}}
\toprule
Word & \multicolumn{1}{c}{green} & \multicolumn{1}{c}{red}   & \multicolumn{1}{c}{n't} & \multicolumn{1}{c}{fight} & \multicolumn{1}{c}{throw} & \multicolumn{1}{c}{hit}   \\ 
\midrule
No. ~(Ori.) & 1 & 2 & 7 & 11 & 18 & 23 \\
PMI (Ori.) & 1.620 & 1.562 & 1.322 & 1.318 & 1.305 & 1.301
\\
No. ~(Mix) & 1 & 2 & 26 & 16 & 52 & 155 \\
PMI (Mix)  & 1.488 & 1.408 & 1.231 & 1.242 & 1.212 & 1.172 \\ 

\bottomrule
\end{tabular}
\caption{PMI calculated for the words in Ori. and Mix datasets. No. is the reverse order in vocabulary according to PMI.} \label{table:pmi}
\end{table}

\begin{figure}[!t]
\centering
\includegraphics[width=77mm]{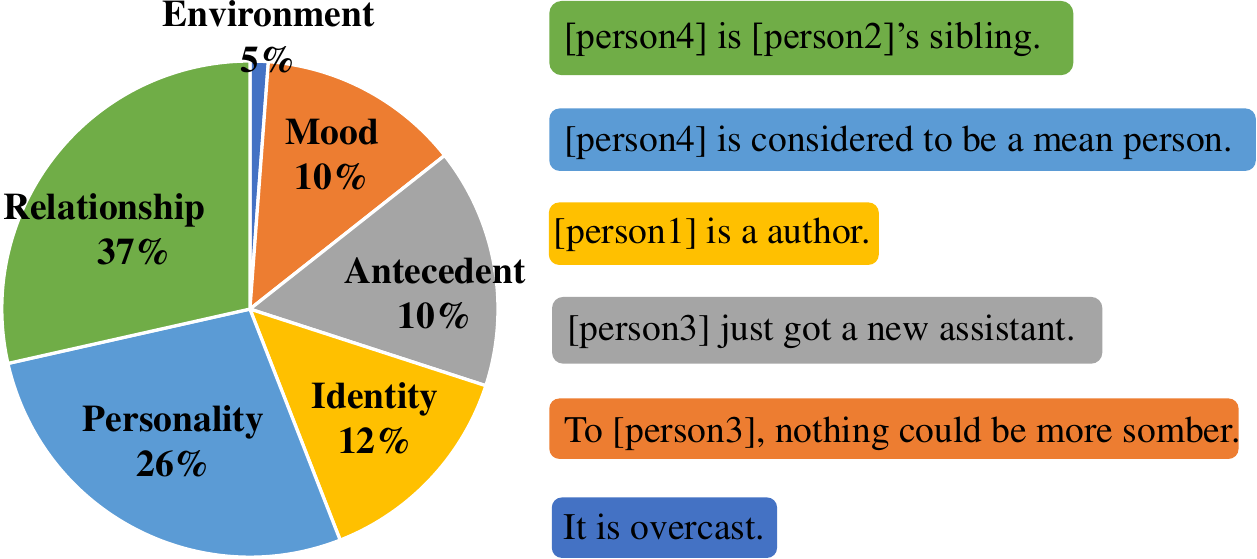}
\caption{An overview of the premises from the samples in validation set.} \label{fig:premise-classification}
\end{figure}

\paragraph{Does Training with Adversarial Samples Help to Relieve the Above Issue?}
Yes, firstly in Table \ref{table:pmi}, we can find that the PMIs of words decrease, indicating that adversarial samples help to balance the correlation of words and labels.

Secondly, Table \ref{table:main-result} shows that there is a great gap in the performance of multi-modal models between testing on OT and AT while training on the original set, which also testifies the existence of bias in the original set. Nevertheless, after training on mixed data, multi-modal models are able to achieve high accuracy both on OT and AT, suggesting that adversarial samples are helpful to improve models' robustness. Besides, We can notice that E-L may be misled by the fewer incorrect samples in the adversarial split, and it results in a decline of 2.8\% on OT, which can be seen as a compromise between generalization and performance.

Thirdly, we find that another huge gap of 39.4\% on OT between BERT trained on the original set and mixed set, and it indicates that the adversarial samples help to change the data distribution and alleviate the bias from the text.

\section{Analysis}
\subsection{Premise Distribution}
From Figure \ref{fig:premise-classification}, we can find that Relationship and Personality account for nearly two thirds of the samples, while the other four categories hold the rest one third, which indicates that the understanding of interpersonal relationships and the correlation between character's personalities and behaviors are test points of PMR. In terms of the reason why there is such a distribution, we maintain that this is because Relationship and Personality are more likely to conform with the instruction of annotation. As presented in Section \ref{task-description}, we expect premises as a supplement of images, so the information implied in premises should not be repeated or contradictory with the content of the image. Choosing the other four premises is more likely to violate the above requirements. For instance, people's mood tends to explicitly show through expression; the dressing of the characters can reflect their identity and occupation.

\subsection{Substitution Strategy}
\begin{table*}[!ht]
\small
\centering
\begin{tabular}{c l p{4.7cm} p{4.7cm}}
\toprule
\textbf{Supertype} & \textbf{Type} & \textbf{Action-True} & \textbf{Distractor1} \\
\midrule
\multirow{19}{*}{PERSON}  & \multirow{3}{*}{Appearance (8.0\%)} & {[}person2{]} with {\color{red}long} hair will   fall on the {[}couch1{]} and have a sleep immediately . & {[}person2{]} with {\color{blue}short} hair will   fall on the {[}couch1{]} and have a sleep immediately . \\ 
\cmidrule{2-4} 
                         & \multirow{3}{*}{Emotion (3.0\%)}    & {[}person1{]} {\color{red}weeps} and tells   {[}person2{]} his sad story , and he listens attentively .                                                 & {[}person1{]} {\color{blue}smiles} and tells   {[}person2{]} his sad story , and he listens attentively .                                                         \\ \cmidrule{2-4} 
                         & \multirow{2}{*}{Clothing (22.0\%)}   & {[}person2{]} will button up his {\color{red}shirt} , because now it does n't look neat .                                       & {[}person2{]} will button up his {\color{blue}coat} , because now it does n't look neat .                                                 \\ \cmidrule{2-4} 
                         & \multirow{2}{*}{Body State (10.0\%)} & As {[}person2{]} is ill {\color{red}in bed} ,   {[}person1{]} will take care of {[}person2{]} .                                                         & As {[}person2{]} is ill {\color{blue}sitting in   the chair} , {[}person1{]} will take care of {[}person2{]} .                                                    \\ 
                         \cmidrule{2-4} 
                         & \multirow{4}{*}{Location (18.0\%)}  & {[}person4{]} {\color{red}who is in a car} will   ask {[}person2{]} to live with him because {[}person2{]} 's roommate has just died of   an accident . & {[}person4{]} {\color{blue} hiding behind the   column} will ask {[}person2{]} to live with him because {[}person2{]} 's roommate has   just died of an accident . \\ 
                         \midrule
\multirow{8}{*}{GENERAL} & \multirow{2}{*}{Objects (23.0\%)}    & {[}person1{]} and {[}person2{]} are   trying to kill each other {\color{red}with knives} .                                                              & {[}person1{]} and {[}person2{]} are   trying to kill each other {\color{blue}with guns} .                                                                         \\ \cmidrule{2-4} 
                         & \multirow{2}{*}{Color (1.0\%) }     & {[}person3{]} with a {\color{red}red} tie will   play a beautiful tune with his guitar .                                                                & {[}person3{]} with a {\color{blue}white} tie will   play a beautiful tune with his guitar .                                                                       \\ \cmidrule{2-4} 
                         & \multirow{2}{*}{Texture (8.0\%)}    & {[}person1{]} will have a rest on the \mbox{{\color{red}chair} .}                                                                                            & {[}person1{]} will have a rest on   the {\color{blue}stone chair} .                                                                                               \\ 
                         \midrule
\multirow{2}{*}{ENVIRONMENT}              & \multirow{2}{*}{Scenery (7.0\%)}    & {[}person1{]} and {[}person2{]} are walking down the {\color{red}{street}} together .                                                                            & {[}person1{]} and {[}person2{]} are walking down the {\color{blue}{park}} together .                                                                                     \\ \bottomrule
\end{tabular}
\caption{Nine types of substitution in Phase 2, the percentage of which in the original validation subset and the examples. For each one, the words before and after the replacement are highlighted in red and blue respectively.}\label{table:substitution-types}
\end{table*}
We manually classified 100 samples from the validation set into nine types, which is presented in Table \ref{table:substitution-types}. As we can see, the types of substitution are varied ,and most of them are related to PERSON supertype, which reflects the emphasis on the characters' behavior.

In terms of the understanding of images, the levels of difficulty vary among different types. For instance, in APPEARANCE, CLOTHING, OBJECTS and COLOR, the words involved in the replacement process are relatively simple, such as long->short, white->red, knives->guns, which are likely to appear as labels in the datasets of object detection. Thus, systems with pre-trained objects detection models are expected to have an excellent performance on the four types. On the contrary, as for the types like EMOTION, BODY STATE and LOCATION, the texts substituted tend to be abstract and mostly are phrases consisting of more than one word. Consequently, these types are more challenging, and in order to succeed on them, the model is supposed to excel at in-depth image understanding and fine-grained grounding. 

Comparing the difference before and after the substitution, we may notice that the substitution, within the scope of several consecutive words, can be regarded as a method of disturbance, and correspondingly, the distractors served as adversarial choices are beneficial to counter the possibility of models exploiting unimodal priors.

\subsection{Case Study}
\begin{figure*}[!ht]
\centering
\subfigure{
\includegraphics[width=0.45\linewidth]{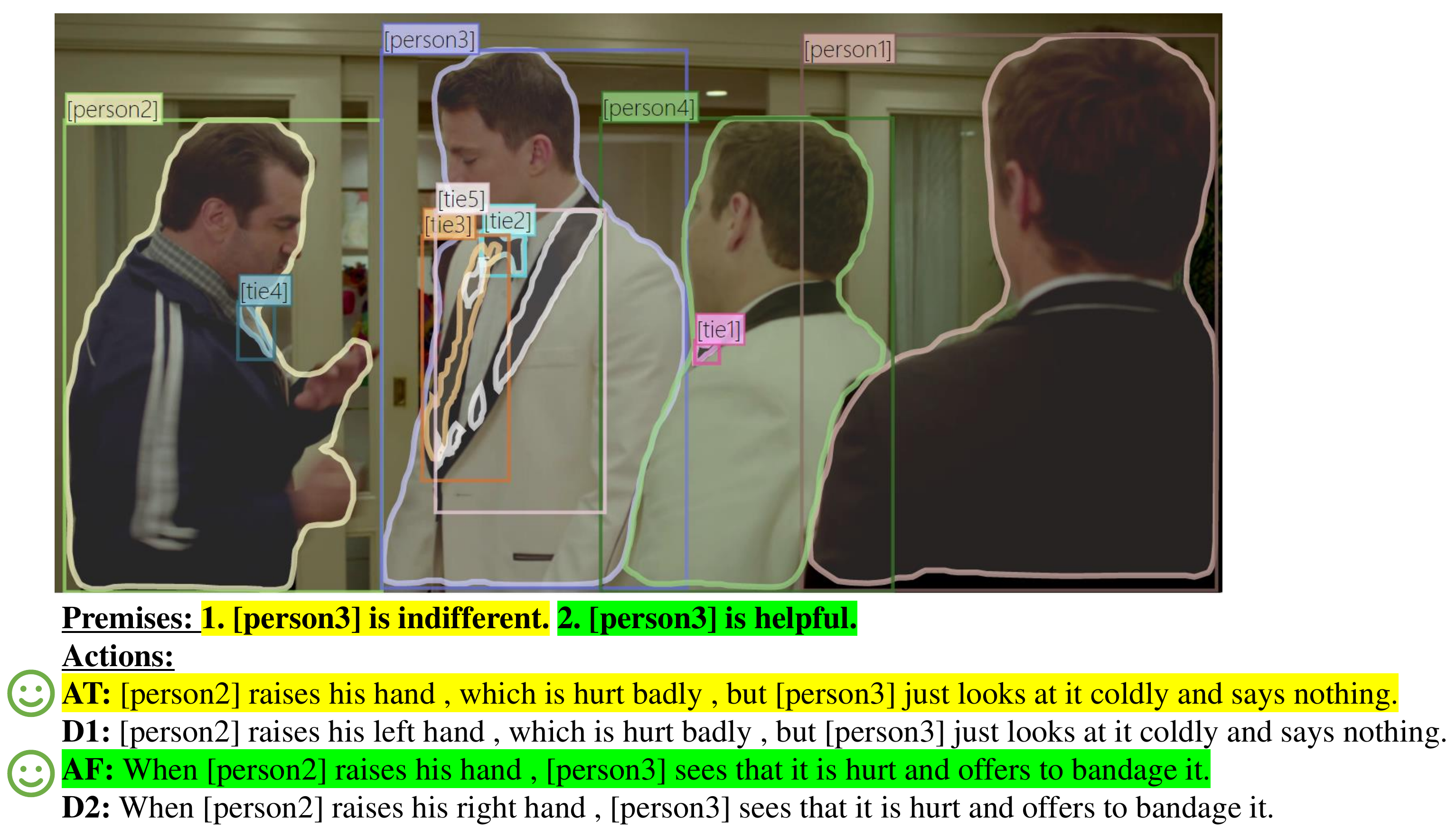}
}
\subfigure{
\includegraphics[width=0.49\linewidth]{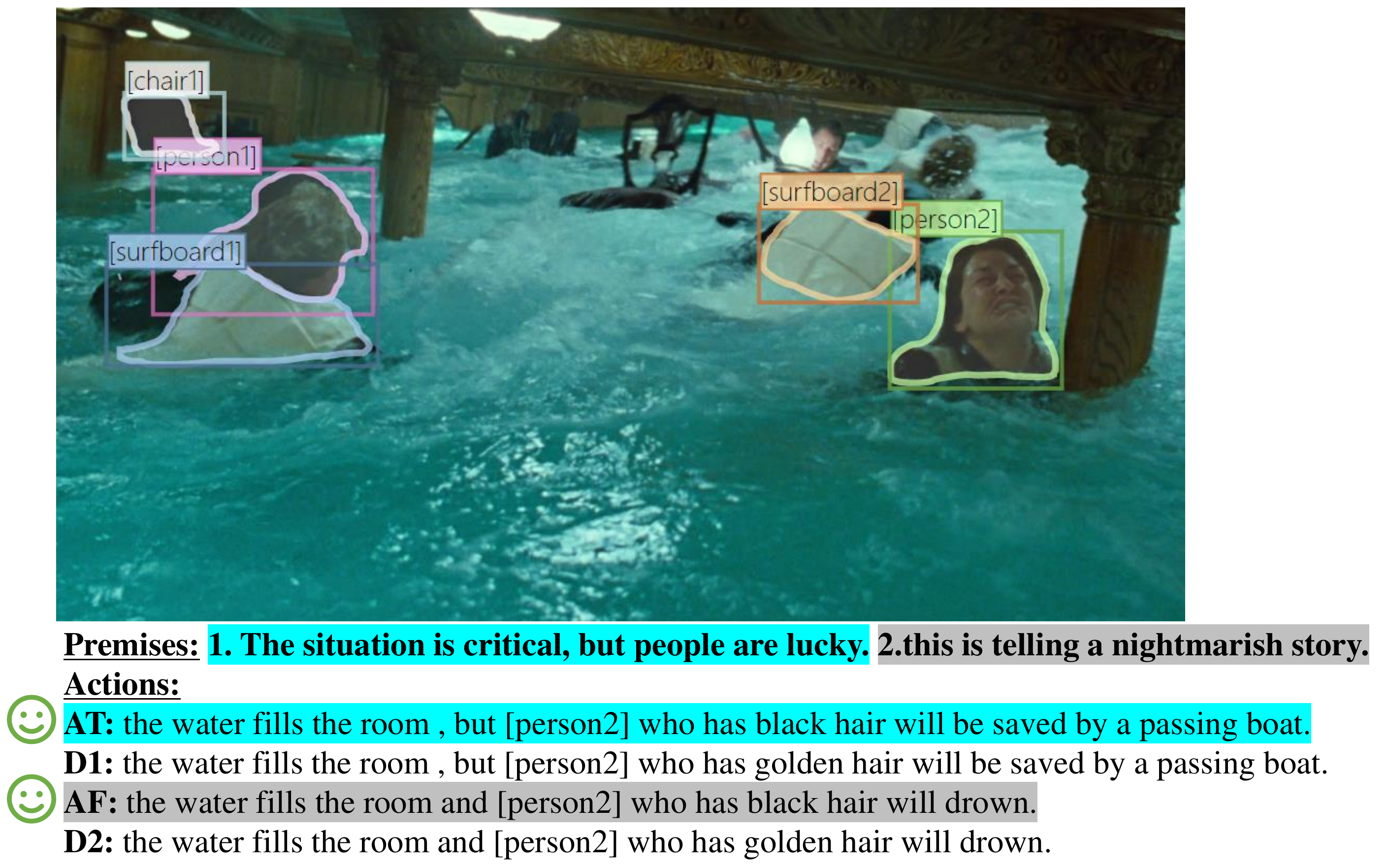}
}
\rule[-1pt]{13cm}{0.05em}
\subfigure{
\includegraphics[width=0.45\linewidth]{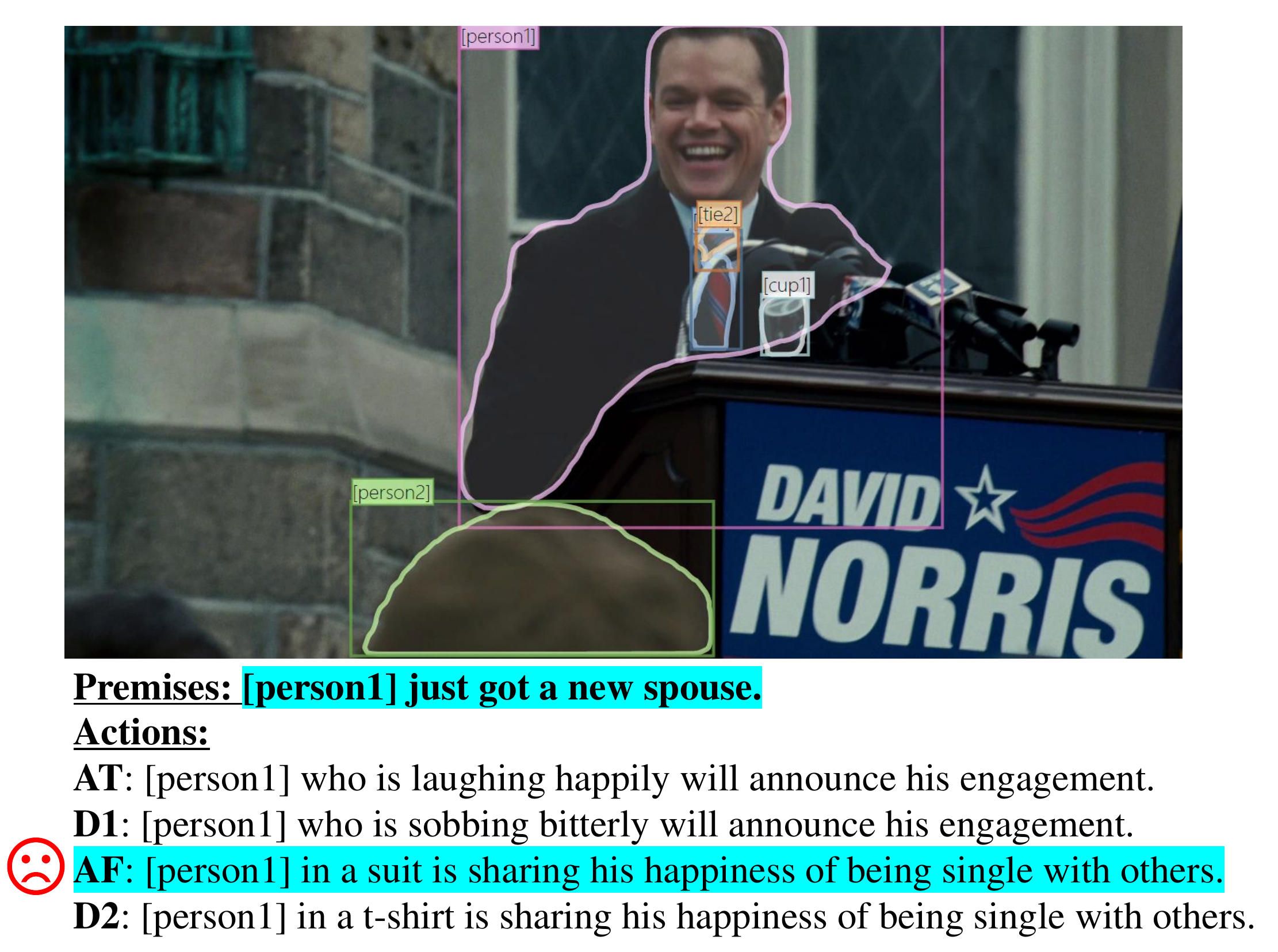}
}
\subfigure{
\includegraphics[width=0.45\linewidth]{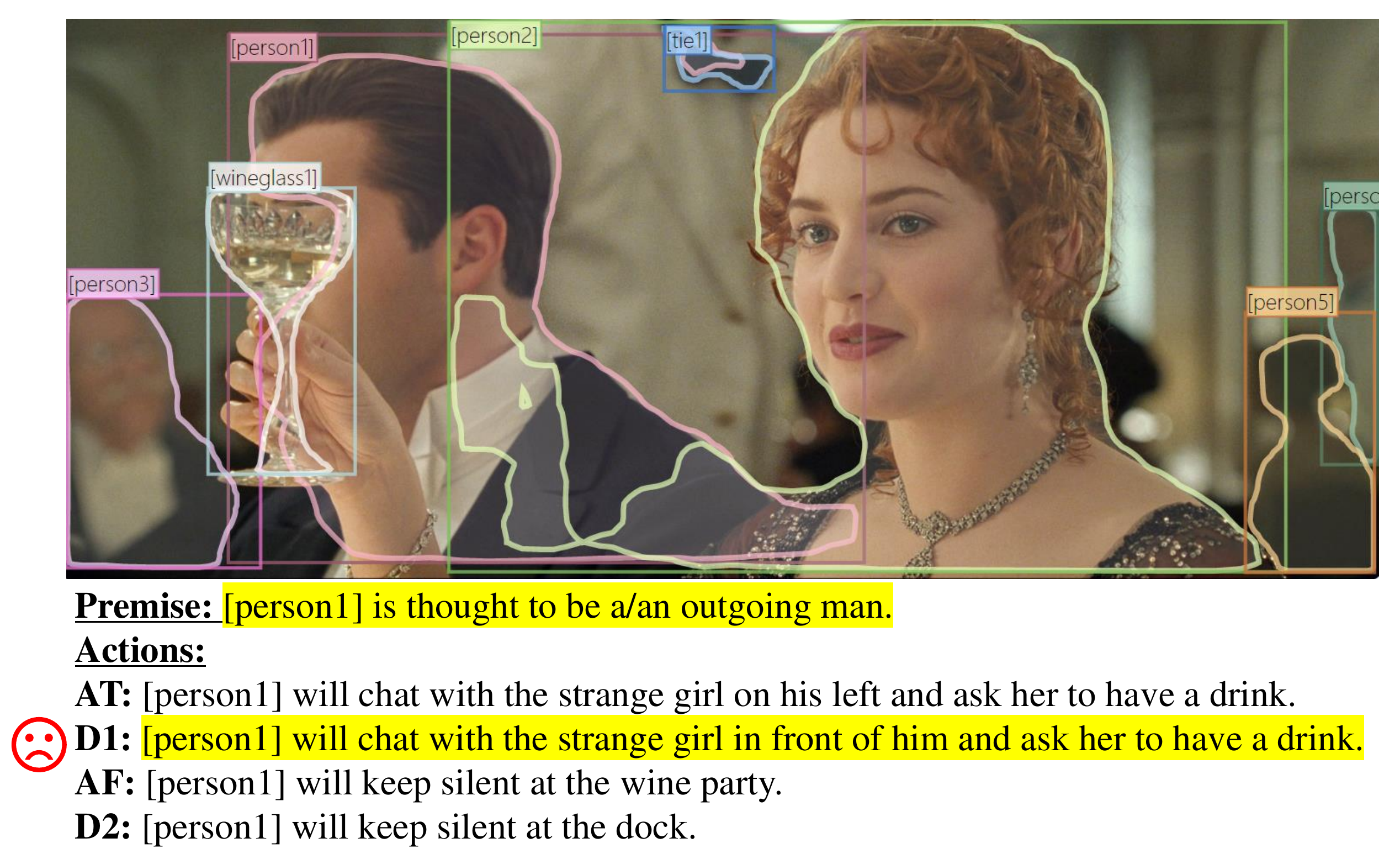}
}
\caption{There are four prediction samples from ERNIE-VIL. For each sample, We list actions in the order of AT, D1, AF and D2, which represent Action-True, Distractor1, Action-False, Distractor2 respectively. Model predictions are highlighted in different colors, which correspond to different premises, and the  predictions are tagged with \smiley{} if correct, and \frownie{} otherwise.}
\label{fig:case-study}
\end{figure*}
We conduct a qualitative analysis shown in Figure \ref{fig:case-study}, where the first two samples are correctly predicted by ERNIE-VIL. For each of them, premise 1 is the original. To explore the influence of different premises on the model predictions with the same image as background,  we write premise 2 for the test, with which the model is expected to choose Action-False as the correct answer. The result shows that ERNIE-VIL predicted correctly both with premise 1 and 2, which indicates the model can distinguish the possible person's behaviors resulting from the two opposite premises. But for case 3, the model makes an incorrect prediction with Action-False chosen, suggesting that the performance is still unsatisfactory due to limited commonsense knowledge. Besides, even if the simple objects substitution can be easily detected by comparing the text with the images directly, case 4 shows that the model is not able to determine the relative position of people, which indicates that a powerful understanding capacity of images is needed to be added.

\section{Related Work}

\paragraph{Multimodal Commonsense Reasoning}
A series of tasks and datasets have been proposed for cross-modal commonsense reasoning.
VCR \citep{vcr} is the most related work to the proposed PMR. It requires machines to understand a image and answer a multi-choice question. Specifically, questions are like `what is going to happen next', `infer the relationship between [personA] and [personB]' and `why is [personA] smiling', as well as the rationale why the answer is true. 
In the video understanding regime, VLEP \citep{lei2020more} is a dataset for future event prediction from videos. Given a video with aligned dialogue, and two possible future events, the AI system is required to choose the more likely event from two provided options.
In the world of cross-modal commonsense graph, VisualCOMET \citep{park2020visualcomet} is a repository of Visual Commonsense Graphs that consists of 1.4 million textual descriptions of visual commonsense inferences. 
The proposed PMR is different from the above-mentioned dataset as we request the models to perceive and understand the source image with the supervision from a specified textual premise, and reason conditionally with joint visual and textual clues.

\paragraph{Natural Language Inference}
Early methods for textual NLI mainly relied on conventional, feature-based methods trained from small-scale datasets \cite{dagan}. The release of larger datasets, such as SNLI \citep{snli:emnlp2015}, MultiNLI \citep{williams2017broad}, made neural network methods feasible. 
In the field of cross-modal entailment, the SNLI-VE \citep{snli-ve} dataset casts the source image as the visual premise, and asks the inference model judge whether the specified textual hypothesis entails or contradicts with the visual premise.
The proposed PMR can also be viewed as a cross-modal entailment task, however difference from SNLI-VE, both premise and hypothetical actions are textual, and the corresponding source images can be viewed as the supplementary evidence while judging the entailment labels. Besides, the existing textual and visual-related NLI tasks are formulated in a three-way classification, while the proposed PMR is model as a 4-choice classification task. 


\section{Conclusion}
We propose a premise-based cross-modal reasoning (PMR) task, in which the textual premise serves as the presumptions to the source images. With selected images and human-curated premise templates, the collected dataset is formulated as a 4-choice classification task, in which we ask human annotators to write one true hypothetical action and three distracting actions. The PMR task gains insights from both the visual commonsense reasoning and the natural language inference tasks. We hope the proposed dataset and task can pave the way for human-like cognitive reasoning from both visual and textual clues.

\section*{Acknowledgements}
This paper is supported by the National Key Research and Development Program of China 2020AAA0106700 and NSFC project U19A2065.

\bibliographystyle{acl_natbib}
\bibliography{anthology}

\clearpage

\appendix

\section*{Appendix}

\section{Author Statement}
Hereby we confirm that we bear all responsibility in case of violation of rights, etc., and confirmation of the data license. This work is licensed under a CC BY-NC license, and both images and annotations can be accessed at https://2030nlp.github.io/PMR/.

\section{Additional Data Analysis}
\begin{figure*}[ht]
    \centering
    \includegraphics[width=120mm]{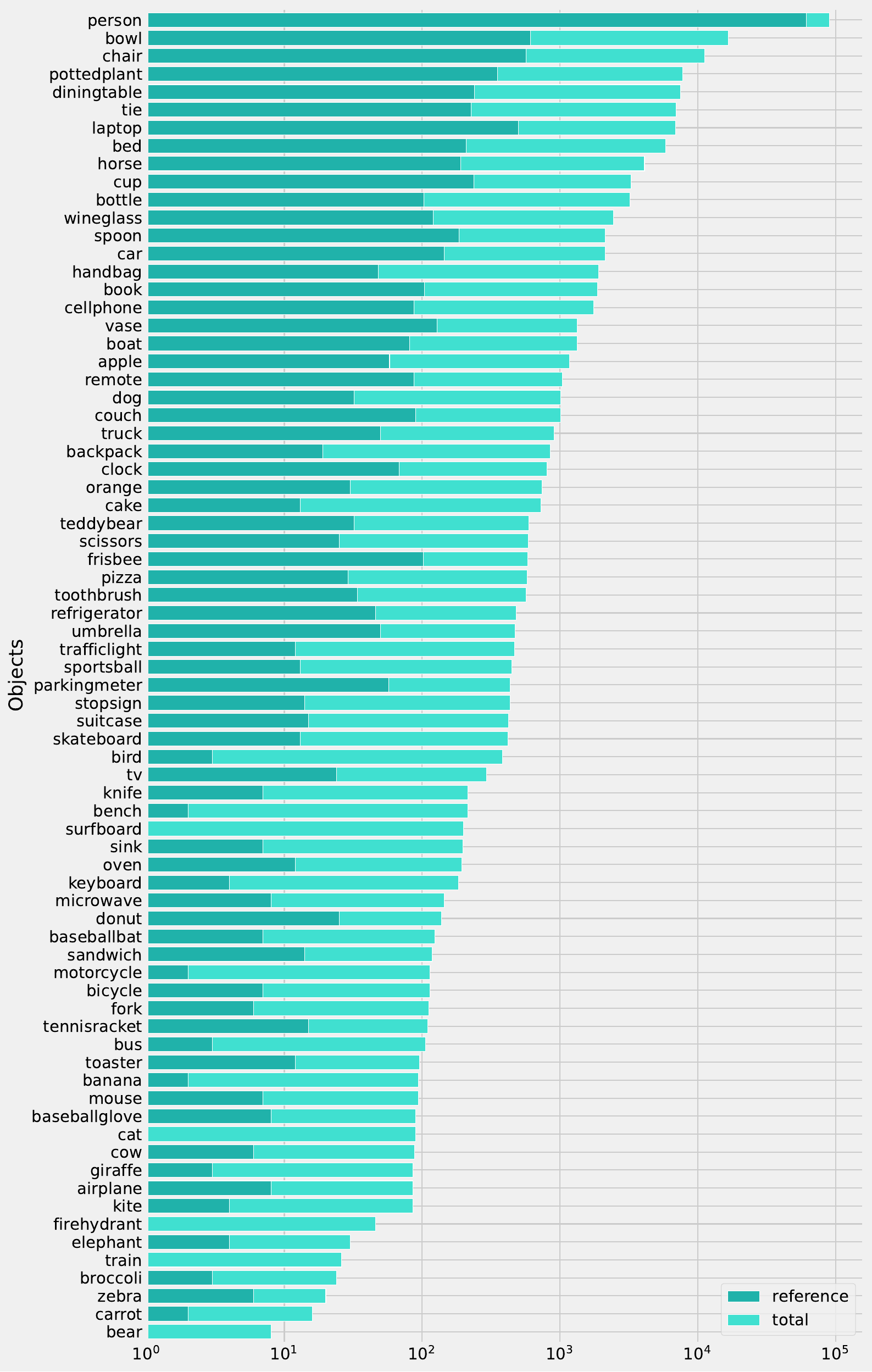}
    \caption{Objects distribution.}
    \label{fig:objects-distribution}	
\end{figure*}

We explore the features of the language in PMR. As is shown in Table \ref{tab:statistics}, the number of 2-grams has reached 84,365, the length of premises average to over 9, and additionally, our action texts average at more than 14 words, which is higher than the common question answering datasets and indicates PMR has high language complexity and diversity.
Furthermore, we calculate the average number of objects mentioned in each sample. Since PMR aims to test the model's capability to infer the action of people, the objects mentioned are mainly persons, but due to our proposed generation strategy of distractors, the text also mention a variety of other objects, such as ‘bowl’ and ‘chair’.  Figure \ref{fig:objects-distribution} shows the objects distribution both in text(reference) and images(total).

\section{Details for Adversarial Samples Generation}
\label{apd-adv}

For each example, manually writing three distractors is expensive and unscalable, which costs nearly 1.5 times as much as a single question-answer pair. Therefore, \citeauthor{vcr} proposes Adversarial Matching to obtain high-quality negatives automatically. For a bunch of question-answer pairs, it aims to take the answer from other question as a negative for a question, with the constraint that the negative is supposed to be relevant to the question, but not overly similar to its true positive at the same time.

Inspired by it, we regarded the 15k premises and their Action-Trues as pairs and took a similar method to double the scale of PMR. Specifically, we chose the most related three actions from other pairs as negatives for a premise according to the relevance scores. Firstly, for a premise $p_i$, in order not to conflict with $o_i$, the objects list of $p_i$'s corresponding image, we performed remapping\footnote{https://groups.google.com/group/visualcommonsense/att\\ach/500d18f416f1b/dataloader.py?part=0.1}, that's substituting the objects mentioned in candidate actions with objects in $o_i$. Secondly, we calculated the relevance scores for $p_i$ with each remapped action by a bert-base model, which was trained with randomly sampled premise action pairs. Then, we picked the top3 actions with the highest confidence as negatives.

This method is simple but effective and efficient compared to the original adversarial matching. Due to the remapping strategy, the objects in actions are going to be randomly substituted, which may cause the contradiction with the premise and thus avoid them being false negatives.
To figure out the effectiveness of our methods, we performed a human test on the 50 items from the adversarial samples generated by it, and the result showed that the human got an accuracy of 0.86 which demonstrated that our methods can generate high-quality negatives.
What's more, overemphasizing the difference with the true positive tends to lead the distractors to be more distinguishable and lower the difficulty of datasets, which was testified by the performance of VL-BERT with an accuracy of 1.0 on the test set of adversarial samples generated in the same way of VCR.

Therefore, we maintain that the remapping strategy is enough to balance the difficulty with the correctness of adversarial samples.

\section{Manually Written Templates for Premises}
\label{premise-templates}
We list all the templates for premises by category. "[ ]" indicates the slots to be filled with words of different types.
\\

\textbf{1. Relationship}
\begin{itemize}
\setlength{\itemsep}{0pt}
\setlength{\parsep}{0pt}
\setlength{\parskip}{0pt}
	\item They are {[}N{]}.
	\item {[}person1{]} and {[}person2{]} are {[}N{]}.
	\item The two speakers are {[}N{]}.
	\item {[}person1{]} is {[}person2{]}'s {[}n{]}.
	\item {[}person1{]} is talking with his/her {[}n{]}.
	\item Because of what happened before, they have a {[}adj{]} relationship now.
	\item The relationship between {[}person1{]} and {[}person2{]} is very {[}adj{]}.
\end{itemize}

\textbf{2. Personality}
\begin{itemize}
\setlength{\itemsep}{0pt}
\setlength{\parsep}{0pt}
\setlength{\parskip}{0pt}
	\item {[}person1{]} is {[}adj{]}.
	\item {[}person1{]} has a {[}adj{]} temper.
	\item Both {[}person1{]} and {[}person2{]} are {[}adj{]}.
	\item {[}person1{]}'s personality is very {[}adj{]}.
	\item {[}person1{]} is thought to be a/an {[}adj{]} man.
	\item {[}person1{]} gives people a {[}adj{]} feeling.
	\item They all know that {[}person1{]} is a/an {[}adj{]} person.
	\item {[}person1{]} is considerd to be a/an {[}adj{]} person.
	\item {[}person1{]} is quite {[}adj{]} about most things.
\end{itemize}

\textbf{3. Identity}
\begin{itemize}
\setlength{\itemsep}{0pt}
\setlength{\parsep}{0pt}
\setlength{\parskip}{0pt}
	\item {[}person1{]} is a {[}job1{]}.
	\item {[}person1{]}'s job is a {[}job1{]}.
	\item The occupation of {[}person1{]} is a {[}job1{]}.
	\item {[}person1{]} works in {[}place1{]}.
	\item {[}person1{]} works for a/an {[}place1{]}.
	\item The line of work that the {[}person1{]} is in is {[}place1{]}
\end{itemize}

\textbf{4. Antecedent}
\begin{itemize}
\setlength{\itemsep}{0pt}
\setlength{\parsep}{0pt}
\setlength{\parskip}{0pt}
	\item There was a/an {[}incident1{]}.
	\item {[}incident1{]} happened.
	\item {[}person1{]} and {[}person2{]} had a fight.
	\item {[}person1{]} had no idea who {[}person2{]} was supposed to be.
	\item {[}person1{]} and {[}person2{]} were very familiar with each other.
	\item {[}person1{]} broke things on accident.
	\item Something unfortunate had just happened.
	\item Something fortunate had just happened.
	\item {[}person1{]} just lost {[}relative1{]}.
	\item {[}person1{]} just got a new {[}relative1{]}.
\end{itemize}

\textbf{5. Mood}
\begin{itemize}
\setlength{\itemsep}{0pt}
\setlength{\parsep}{0pt}
\setlength{\parskip}{0pt}
	\item {[}person1{]} is {[}adj{]}.
	\item {[}person1{]} feels {[}adj{]}.
	\item To {[}person1{]}, nothing could be more {[}adj{]}.
	\item Having a conversation with {[}person2{]}, {[}person1{]} is very {[}adj{]}.
	\item {[}person1{]} has a {[}adj{]} time with {[}person2{]}.
	\item Because of {[}person2{]}'s behavior, {[}person1{]} feels very {[}adj{]}.
	\item {[}person1{]} is so {[}adj{]} to talk with {[}person2{]}.
	\item people have {[}adj{]} looks on their faces.
	\item {[}person1{]} looks {[}adj{]} today.
	\item Recently, {[}person1{]} is becoming more and more {[}adj{]}.
	\item This scene makes them very {[}adj{]}.
	\item {[}person1{]} makes {[}person2{]} {[}adj{]}.
	\item {[}person1{]} is feeling a bit {[}adj{]}.
	\item {[}person1{]} is in a {[}adj{]} mood today.
	\item {[}person1{]} is filled with {[}n{]}.
\end{itemize}
\textbf{6. Environment}
\begin{itemize}
\setlength{\itemsep}{0pt}
\setlength{\parsep}{0pt}
\setlength{\parskip}{0pt}
	\item It is {[}weather1{]}.
	\item The atmosphere is {[}emotion1{]}.
	\item There is a/an {[}emotion1{]} in the air.
	\item This is telling a {[}emotion1{]} story.
	\item The scene is {[}emotion1{]}.
\end{itemize}

\section{Payment for Worker}
\label{pricing-strategy}
Crowd workers performed annotations mainly in phase1 and phase2, and we paid them different prices according to the workload and the quality. In Phase 1, workers are responsible for selecting the premise and writing two hypothetical actions. If accepted in Phase 2, they would be paid 0.15\$ for each sample. And they would only obtain 1.5 cents per sample if rejected. In Phase 2, the total salary was calculated by multiplying the basic salary by qualification rate. For basic salary, if anyone accepted the annotation in Phase 1 and generate another two distractors, he or she would get 7.5 cents, otherwise, they only got 1.5 cents (No distracter needs to be generated in such a case certainly). In terms of qualification rate, it was obtained by checking the quality of the result in Phase 2, and we would calculate it for every worker.
On average, the annotators were paid about 3 times as much as the prevailing local minimum wage per hour.

\section{Data Availability and Copyright}\label{copyright}
According to Section 107 of the Copyright Law\footnote{https://www.copyright.gov/title17/ 92chap1.html107},and 28A and 30 of the Copyright Acts\footnote{https://www.gov.uk/government/publications/copyright-acts-and-related-laws}, there is one exception to copyright infringement which is fair use (or fair dealing). Fair use is appropriate for public benefit purposes, like research.  Our use is not of commercial nature.   Besides,  we only use texts  that  are  publicly  available,  and  the  source will be stated according to law. Users can download the images directly from the original source.

\section{Experimental Details}
We list all the hyperparameters training different models in Table \ref{hyperparameters}.

\begin{table*}[!ht]
\begin{tabular}{l|llllll}
\toprule
                & BS & CPU/GPU                   & LR & Optimizer & Warmup Steps           & Epochs/Steps \\ \midrule
BERT            & 16         & CPU                       & 5e-5          & Adam      & 500                    & 10 epochs          \\
VL-BERT-Base    & 4          & 4 2080T GPUs & 7e-5          & SGD       & 1000                   & 20 epochs          \\
ERNIE-VIL-Base  & 4          & 1 2080T GPUs & 2e-5          & Adam      & 8000                   & 22500 steps        \\
UNITER-Base     & 16         & 1 2080T GPUs & 3e-5          & AdamW     & 1084 & 20 epochs          \\
VL-BERT-Large   & 4          & 4 A40 GPUs                & 7e-5          & SGD       & 1000                   & 20 epochs          \\
ERNIE-VIL-Large & 4          & 1 A40 GPUs                & 2e-5          & Adam      & 8000                   & 22500 steps        \\
UNITER-Large    & 16         & 1 A40 GPUs                & 3e-5          & AdamW     & 1084 & 20 epochs         \\
\bottomrule
\end{tabular}
\caption{Hyperparameters for training.}\label{experimental-details} \label{hyperparameters}
\end{table*}

\section{Annotation Interface}
\label{annotation-interface}
\begin{figure*}[ht]
    \centering
    \includegraphics[width=120mm]{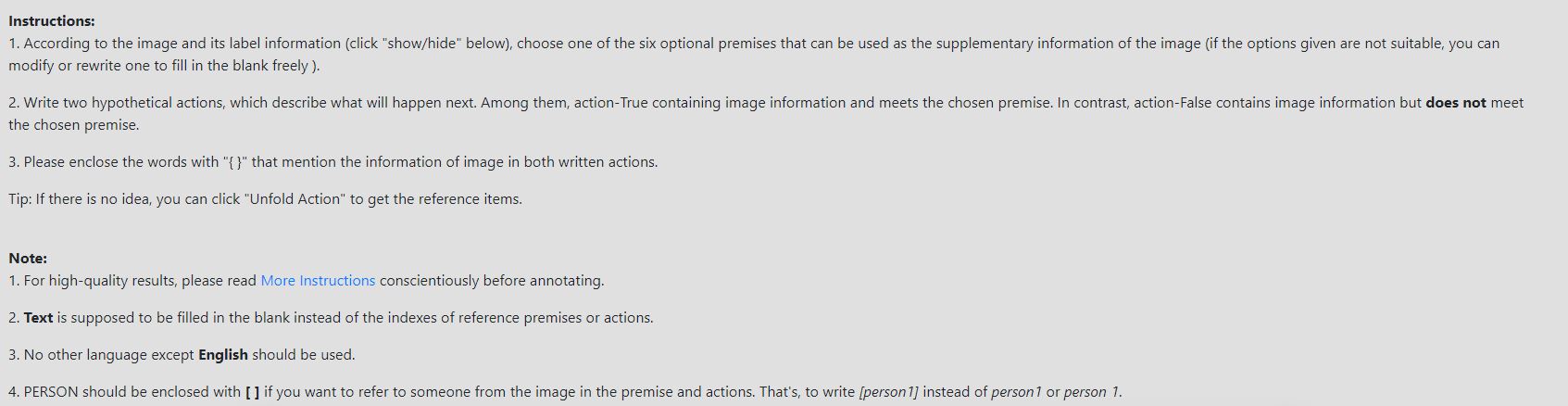}
    \caption{Annotation Instruction.}
    \label{fig:instruction}	
\end{figure*}

\begin{figure*}[ht]
    \centering
    \includegraphics[width=120mm]{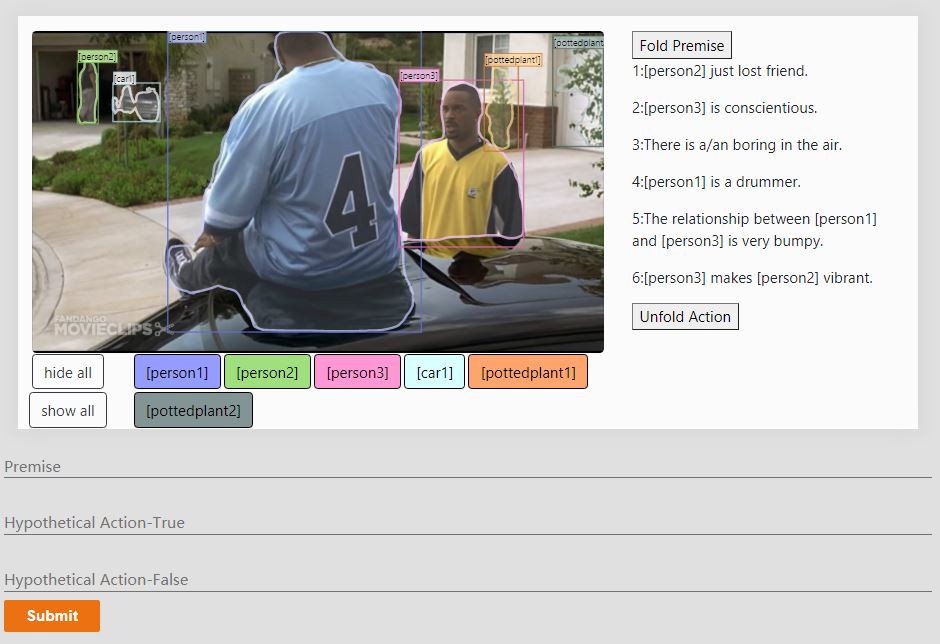}
    \caption{Annotation interface for phase 1.}
    \label{fig:phase1}

\end{figure*}

\begin{figure*}[ht]
    \centering
    \includegraphics[width=120mm]{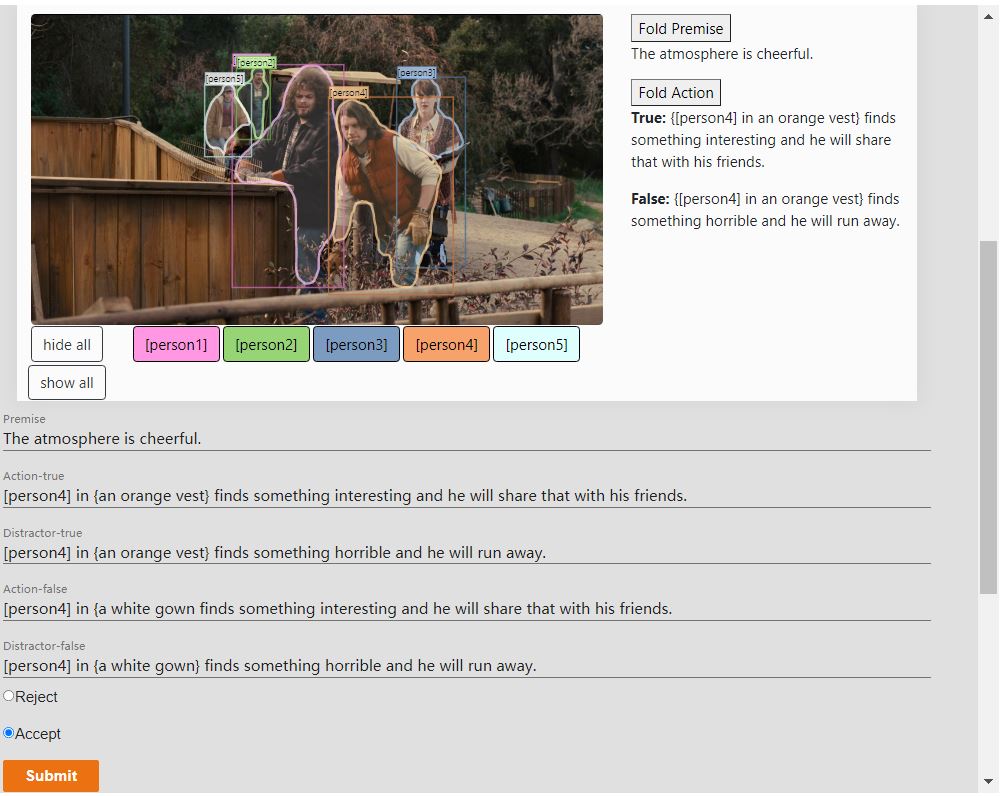}
    \caption{Annotation interface for phase 2.}
    \label{fig:phase2}
\end{figure*}

\begin{figure*}[t]
    \centering
    \includegraphics[width=120mm]{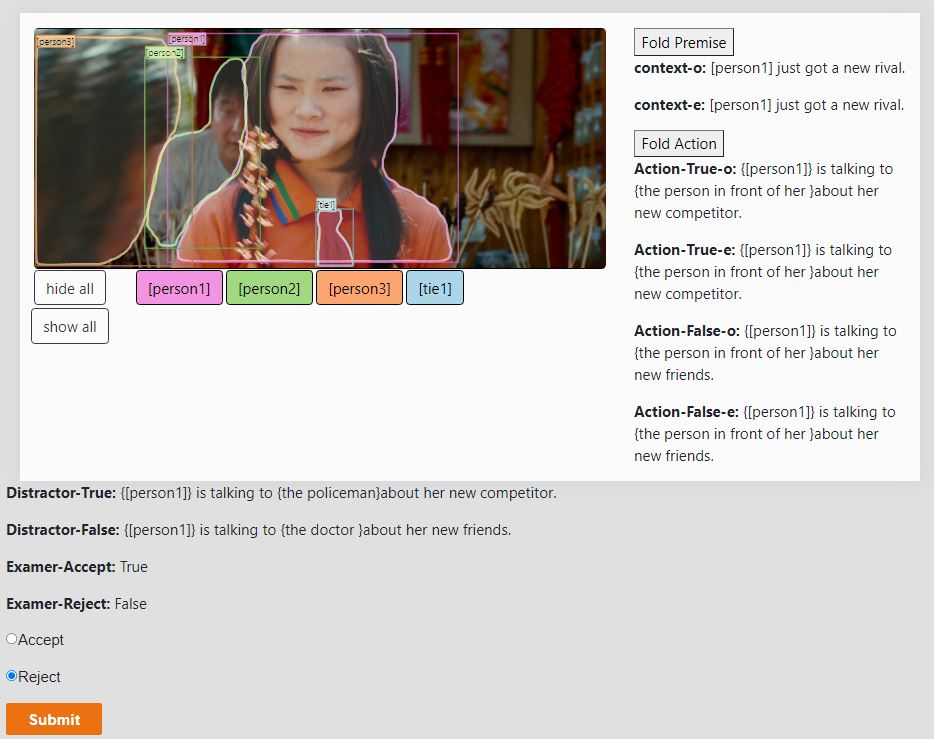}
    \caption{Interface for quality control.}
    \label{fig:phase3}
\end{figure*}

We present instruction for annotation in Figure \ref{fig:instruction}, and the interfaces\footnote{The interfaces is constructed refering to the code from https://visualcommonsense.com/explore} used in 3 phases in Figure \ref{fig:phase1}, Figure \ref{fig:phase2} and Figure \ref{fig:phase3}.

\end{document}